%% file: main-paperID.tex
\documentclass[runningheads]{llncs}

\usepackage{graphicx}
\usepackage{comment}
\usepackage{amsmath,amssymb}
\usepackage{multirow}
\usepackage{colortbl}
\usepackage{tabularx}
\usepackage{booktabs}
\usepackage{color}
\usepackage{url}
\usepackage{hyperref}
\usepackage[numbers]{natbib}
\usepackage{cleveref}
\usepackage{overpic}

\usepackage{caption}
\usepackage{subcaption}
\usepackage[dvipsnames,table]{xcolor}

\newif\ifreview
\reviewfalse

\ifreview
	\usepackage{lineno}

	\linenumbers
\fi

\begin{document}

%%%%%%%%%%%%%%%%%%%%% Add submission id, track, and title. %%%%%%%%%%%%%%%%%%%%%

% TODO: Please insert your submission number here
\def\SubNumber{4}

% TODO: Please uncomment the track this paper will be submitted to, comment all other lines
\def\GCPRTrack{Main Track}
% \def\GCPRTrack{Special Track: Pattern recognition in the life and natural sciences}
%\def\GCPRTrack{Special Track: Photogrammetry and remote sensing}
%\def\GCPRTrack{Young Researcher's Forum}
%\def\GCPRTrack{Fast Review Track}

% TODO: Replace with your title
\title{Gaussian Splatting in Style}
% You can use \thanks for acknowledgment. Do not add any acknowledgment to the draft 
% version that is used for the review process.  
% \title{Title\thanks{XXX}}

\ifreview
	% ANONYMOUS SUBMISSION FOR REVIEW
	% DO NOT MODIFY these for the draft version that is used for the review process.
	\titlerunning{GCPR 2024 Submission \SubNumber{}. CONFIDENTIAL REVIEW COPY.}
	\authorrunning{GCPR 2024 Submission \SubNumber{}. CONFIDENTIAL REVIEW COPY.}
	\author{GCPR 2024 - \GCPRTrack{}}
	\institute{Paper ID \SubNumber}
\else
 %\orcidID{0000-1111-2222-3333}

	\author{Abhishek Saroha\inst{1,2} \and
	Mariia Gladkova\inst{1,2} \and
	Cecilia Curreli\inst{1,2} \and
        Dominik Muhle \inst{1,2} \and
        Tarun Yenamandra\inst{1,2} \and
        Daniel Cremers\inst{1,2}}

	\authorrunning{A. Saroha et al.}
	
	\institute{Technical University of Munich \and Munich Center for Machine Learning, Germany}
	% \url{http://www.springer.com/gp/computer-science/lncs} \and ABC Institute, Rupert-Karls-University Heidelberg, Heidelberg, Germany\\
 
\fi

\maketitle              % typeset the header of the contribution

\begin{figure}
% \begin{center}
\centering
\vspace{-19pt}
\includegraphics[width=0.9\textwidth]{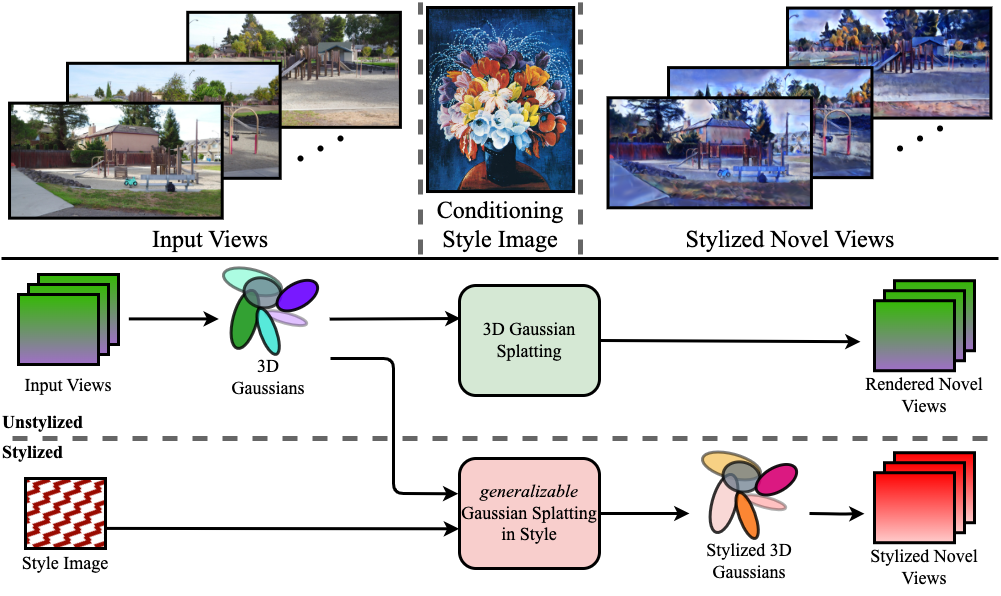}
\vspace{-5pt}
\caption{
% Given multi-view images of a real-world scene, we perform the task of scene stylization. We train a network that, provided with style image at test time, generates stylized novel views of the scene conditioned on the input style in real-time that are consistent in 3D space. In contrary to popular scene stylization approaches of fitting a scene with each new style, we learn this mapping via a neural network and therefore, can generate novel views of the scene even for unseen styles at a rate of roughly 150FPS.
Given multi-view images of a real-world scene, we perform the task of scene stylization. Unlike previous scene stylization approaches, we do not need to fit a scene to each new style. Employing a neural network, conditioned on a style image, allows us to generalize to a variety of styles. Our method can generate 3D consistent scene stylization at approximately $150$ FPS.
% We train a network that, provided with style image at test time, generates stylized novel views of the scene conditioned on the input style in real-time that are consistent in 3D space. In contrary to popular scene stylization approaches of fitting a scene with each new style, we learn this mapping via a neural network and therefore, can generate novel views of the scene even for unseen styles at a rate of roughly 150FPS.
}
\label{fig:teaser}
\vspace{-1cm}
% \end{center}
\end{figure}

\begin{abstract}
3D scene stylization extends the work of neural style transfer to 3D. A vital challenge in this problem is to maintain the uniformity of the stylized appearance across multiple views. A vast majority of the previous works achieve this by training a 3D model for every stylized image and a set of multi-view images. In contrast, we propose a novel architecture trained on a collection of style images that, at test time, produces real time high-quality stylized novel views. We choose the underlying 3D scene representation for our model as 3D Gaussian splatting. We take the 3D Gaussians and process them using a multi-resolution hash grid and a tiny MLP to obtain stylized views. The MLP is conditioned on different style codes for generalization to different styles during test time. The explicit nature of 3D Gaussians gives us inherent advantages over NeRF-based methods, including geometric consistency and a fast training and rendering regime. This enables our method to be useful for various practical use cases, such as augmented or virtual reality. We demonstrate that our method achieves state-of-the-art performance with superior visual quality on various indoor and outdoor real-world data. 

\end{abstract}
\input{sections/intro}

% \vspace{-0.7cm}
\input{sections/related_work}
% \vspace{-0.6cm}
\input{sections/preliminaries}

\input{sections/method}
\input{sections/results}

% \vspace{-0.6cm}
\input{sections/ablations}
% \vspace{-0.6cm}
\input{sections/conclusion}

\bibliographystyle{splncs04}
\bibliography{egbib}

\end{document}

%%%%%%%%%%%%%%%%%%%%% Add submission id, track, and title. %%%%%%%%%%%%%%%%%%%%%

% TODO: Please insert your submission number here
\def\SubNumber{4}

% TODO: Please uncomment the track this paper will be submitted to, comment all other lines
\def\GCPRTrack{Main Track}
% \def\GCPRTrack{Special Track: Pattern recognition in the life and natural sciences}
%\def\GCPRTrack{Special Track: Photogrammetry and remote sensing}
%\def\GCPRTrack{Young Researcher's Forum}
%\def\GCPRTrack{Fast Review Track}

% TODO: Replace with your title
\title{Gaussian Splatting in Style}
% You can use \thanks for acknowledgment. Do not add any acknowledgment to the draft 
% version that is used for the review process.  
% \title{Title\thanks{XXX}}

\appendix

\section{Qualitative Scene Video Results}
\label{sec:videos}
We provide videos of the stylized scenes to demonstrate the superior performance of our proposed method against the baselines. It can be observed that our method, not only maintains geometric consistency, but is also able to successfully able to transfer the required style information to the rendered views. 

\section{Additional Comparisons with ARF}
\begin{figure}
    \centering
    \includegraphics[width=0.8\linewidth]{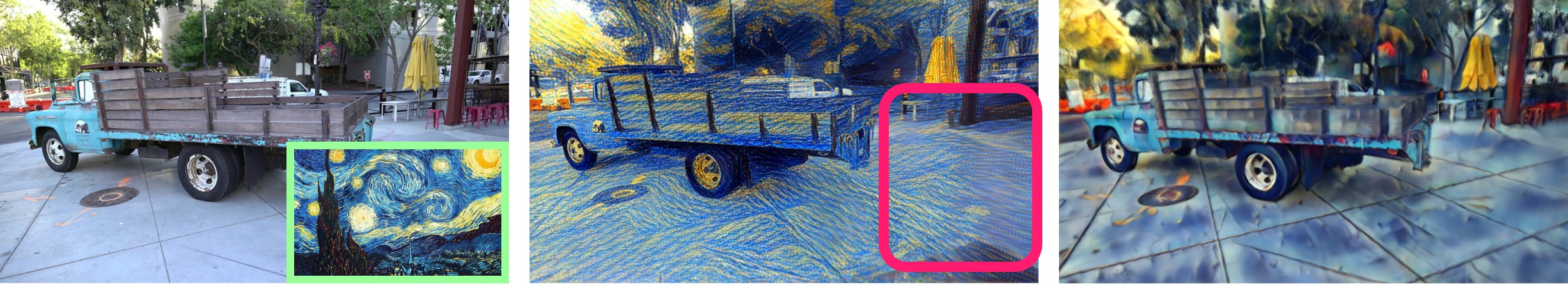}
    \caption{Here we show additional qualitative ablations with ARF\cite{zhang2022arf}. We find that despite its aggressive texture transfer abilities, ARF fails to create consistent texture across the entire surface and totally misses stylizing on far away objects and backgrounds.  GSS on the other hand, is able to perform stylization uniformly and consistently for distant objects and background such as the floor or buildings. We refer the reader to \Cref{sec:videos} for more evidence to support our comparison.}
    \label{fig:arf_comparison}
\end{figure}

We further qualitatively compare our method against ARF\cite{zhang2022arf}. As explored in the main text, ARF has great ability to perform texture transfer, however it suffers from stylizing in the entire space, especially in open outdoor scenes such as the truck scene explored in \Cref{fig:arf_comparison}. The stylization is inconsistent for instance, on the floor or non existing on far away objects such as the background trees or the sky. On the contrary, our method is capable of stylizing far distant objects including the sky while maintaining consistency on large areas such as the ground, while being faithful to the original geometry. It is also worth acknowledging that unlike ARF, we do not need to retrain the scene for each new style, thus making our method more versatile. 

\bibliographystyle{splncs04}
\bibliography{egbib}

%% file: sections/intro.tex
\begin{minipage}{\linewidth}
\begin{center}
\includegraphics[width=0.79\linewidth]{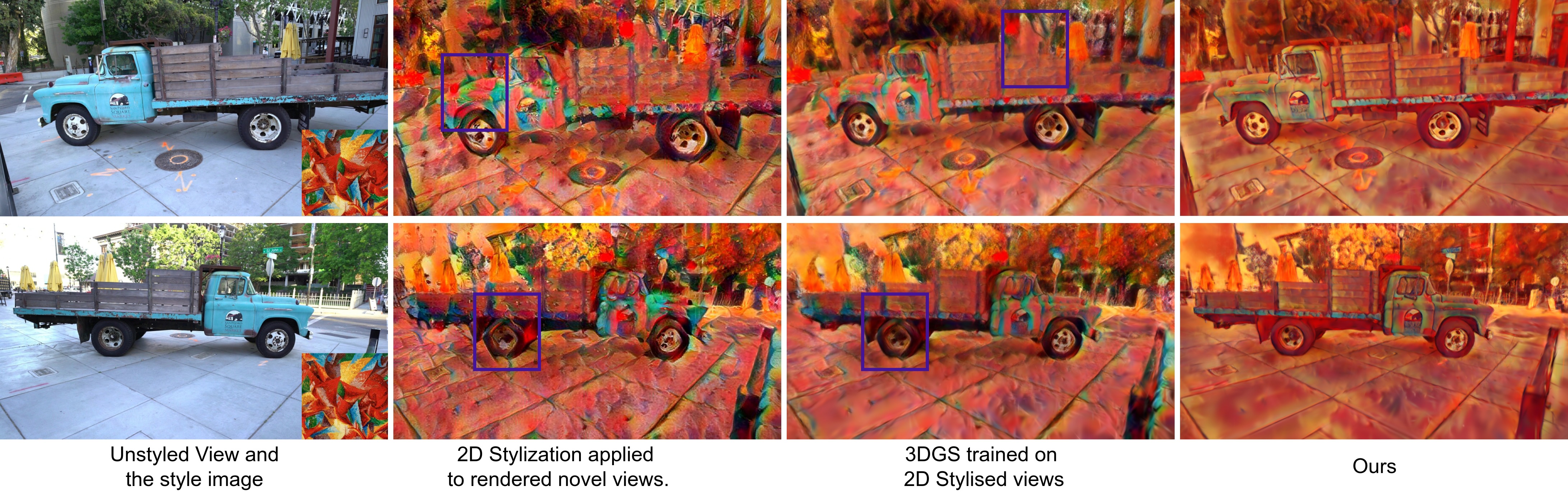}
\vspace{-0.25cm}
\captionsetup{hypcap=false}
\captionof{figure}{The motivation from our work stems from the requirement of a specialized method that, while stylizing a scene, considers the spatial information into account. We show that to generate stylized novel views of a scene, it is insufficient to stylize the rendered views or train a scene representation model on stylized 2D images. It leads to loss of information, such as deformity in the solid truck's body shown above.}
\end{center}
\label{fig:intro_fig}
% \vspace{-0.85cm}
\end{minipage}
\section{Introduction}
\label{sec:intro}
\vspace{-0.2cm}
% \begin{tikzpicture}[spy using outlines={rectangle, magnification=2, size=1cm, connect spies}]
%     \node[anchor=south west,inner sep=0] (image) at (-0.5,0) 
%     {\includegraphics[width=\linewidth]{images/armadillo/64/oursno_color.png}};
%     \begin{scope}[x={(image.south west)},y={(image.north east)}]
%         % \draw[red,thick] (0.5,0.5) rectangle (0.6,0.6); % Adjust coordinates as necessary
%         \spy[red] on (0.8, 0.5) in node (image) [left] at (0.3,0.5); % Adjust coordinates as necessary
%     \end{scope}
%     \end{tikzpicture}

% \vspace{-0.25cm}

Driven by the curiosity of how a particular image would appear if painted by Van Gogh, Picasso and other artists, style transfer has been at the heart of the image processing community for a long time. It can be seen as a challenge of texture transfer.  The core idea of neural style transfer is, given a style image with desired aesthetics and a target image with pursued content, to apply the features of a style image such as colors and texture to the content image while keeping information inside the content image intact. Earlier works \cite{efros1999texture,wei2000fast,kwatra2003graphcut, efros2023image} devised powerful algorithms for texture synthesis. Further advancements in deep learning gave rise to works \cite{gatys2016image,johnson2016perceptual,huang2017adain} that approach the problem by suitably trained neural networks. The ability of neural networks to effectively grasp the features of an image at multiple levels of details made them a preferred choice over non-learning based algorithms\cite{gatys2016image}. 
Fuelled by the availability of 3D data, this phenomenon of style transfer was extended to the stylization of scenes\cite{huang2022stylizednerf,huang2021learning,liu2023stylerf,zhang2022arf}, and is referred to as {\em scene stylization}, which is the primary focus of this work. 

%In this work, we propose a solution for the task of scene stylization given a style image. 
The ability  to control the visual properties of a scene is central to many practical applications such as avatar modeling or scene editing \cite{pang2023ash,moreau2023human,qian2023gaussianavatars,zheng2023gpsgaussian} to name a few. With the wide adaptability and increasing interest in various augmented and virtual reality applications, it is not only necessary that these techniques are accurate, but also real-time capable to ensure the best user experience. 

Over the past few years, different methods have been proposed to efficiently represent a complex 3D scene. \cite{niklaus20193d,wiles2020synsin,tulsiani2018factoring,shin20193d} were some methods leveraging an explicit pointcloud representation. The domain witnessed an incredible pace of growth following the pioneering work of Neural Radiance Fields(NeRFs)\cite{mildenhall2021nerf, mildenhall2019local}. NeRFs gained large popularity, primarily due to its ability to represent a complex scene in the weights of a single multi-layer perceptron(MLP), while capturing fine details during novel view synthesis. Nonetheless, the proposed method required long training time and it was slow during inference. Therefore, many variants of NeRFs emerged such as \cite{reiser2021kilonerf,yu2021pixelnerf,barron2021mipnerf,muller2022instant,niemeyer2021regnerf,Chen2022tensorf} that improved the seminal work significantly. Recently, 3D Gaussian Splatting(3DGS)\cite{kerbl3Dgaussians} further pushed the boundaries of novel view synthesis, especially in terms of consistency and speed. 3DGS is, by design, more explicit in terms of storing scene information in the form of 3D Gaussians, with each Gaussian having certain properties attached to it. Additionally, 3DGS renders views in real-time, thus making it highly suitable for practical applications.  

In this paper, we tackle the problem of learning to stylize a set of 3D Gaussians given a set of style images, namely scene stylization. Therefore, we call our approach as Gaussian Splatting in Style(GSS). We condition this representation on a style image to obtain stylized views of the complex 3D scene. Utilizing these Gaussians as the backbone of our approach ensures spatial consistency. Due to the proposed design, our model stylizes a scene given any style image and does not require re-training as some of the previous works on scene stylization such as ARF~\cite{zhang2022arf}. Furthermore, our approach does not add any overhead to the existing real-time rendering speed of 3DGS, making it a preferred choice for AR/VR applications. We demonstrate the superiority of our method on well-established consistency metrics, showing the best performance in both short-term and long-term measurements. We further support our claims with the qualitative analysis of stylized rendered views comparing our work to existing scene stylization baselines.

To summarise, our contribution in this paper are as follows:
\begin{itemize}
    \item We develop a novel state-of-the-art method, GSS, to perform neural scene stylization in real-time based on 3D Gaussian splatting. We are among the first to perform scene stylization using 3D Gaussians. 
    \item We demonstrate the effectiveness of our method by comparing against various types of baselines both quantitatively and qualitatively across various real-world datasets across different settings.
\end{itemize}

%% file: sections/related_work.tex
\section{Related Work}
\label{sec:related}
%\vspace{-0.35cm}
\subsection{3D Scene representation}
Complex 3D scenes can be represented in many ways. One of the most common and well known ways is that of using point clouds\cite{huang2021learning,cao2020psnet}. Similar to point clouds, some works such as \cite{schwarz2022voxgraf,yu2021plenoxels} use a voxel grid for representing a 3D scene. Voxel grids are not the most ideal choice for scene representation due to their high memory footprint. Neural implicit scene based methods, led by the works of \cite{park2019deepsdf,Occupancy_Networks,Peng2020convonet} making use of signed distance fields(SDFs) and occupancies respectively, helped solve this memory issue. In principle, implicit neural representations can generate meshes and surfaces upto an arbitrarily high resolution. Recently, work of NeuralAngelo \cite{li2023neuralangelo} set a new benchmark as it could recover a highly detailed 3D surface of a large-scale scene from a set of input images. Following a similar direction neural radiance fields, also known as NeRFs\cite{mildenhall2021nerf} are also an implicit method that is described in further detail below.
% \Cref{sec:nvs}. 
\vspace{-0.4cm}
\subsection{Novel View Synthesis}
\label{sec:nvs}
The task of generating unseen views of an object/scene given a collection of input images is known as novel view synthesis. Classical works, such as \cite{Levoy1996Light,Chen1993View,Debevec1996Modeling,Zitnick2004High,Heigl1999Plenoptic,Seitz1996ViewMorphing,Gortler1996Lumigraph} aimed at generating novel views directly from the given set of multi-view images. With the onset of deep learning, new approaches such as \cite{Hedman2018Deep,Sitzmann2019Scene,Sitzmann2019Deepvoxels,riegler2020free} paved the way for neural network-based novel view synthesis(NVS). All these methods represented the scene in their own ways. A major milestone in the domain of NVS was with the advent of Neural Radiance Fields(NeRFs) \cite{mildenhall2021nerf}. NeRFs represent the scene as the weights of a multi-layer perceptron(MLP). The input to these MLPs are points in space and the viewing direction, which then gives out the density and color for each queried point. Despite its success, NeRF suffered from many drawbacks. A slow training regime, followed by the need to optimize it for each scene made it less viable for many use cases. It was also observed that NeRFs performed extremely well when the input views were dense and covered many angles of the scene. It often failed to get the finer details from views far-away from those used in training. Follow up works such as \cite{niemeyer2021regnerf,chibane2021stereo,wang2023sparsenerf,yu2021pixelnerf} focused specifically on solving NVS for a sparse set of input views. Meanwhile, \cite{yu2021pixelnerf,wimbauer2023bts,nguyen2023cascaded} focused on a more generalized framework to avoid fitting a MLP to each new scene, while \cite{barron2021mipnerf,yu2021plenoxels,reiser2021kilonerf,muller2022instant} improved on the original work in terms of training and inference speed. 
Recently, 3D Gaussian Splatting(3DGS) \cite{kerbl3Dgaussians} also tackled the task of novel view synthesis in real-time. We explain the idea of 3DGS in \Cref{sec:prelim} and can recommend the readers to visit \cite{kerbl3Dgaussians} for more details. 

\vspace{-0.55cm}
\subsection{Style Transfer and Scene Stylization}
\vspace{-0.25cm}
Neural style transfer is the task of transferring the style of a source image onto the target(or content) image\cite{gatys2016image}. It is a complex task as the transfer needs to preserve the information present in the target image. This problem can also be formulated as that of texture transfer, and was tackled in the works of \cite{efros1999texture, efros2023image,wei2000fast,kwatra2003graphcut} before the widespread adoption of neural networks. Later, the pioneering work of \cite{gatys2016image} successfully applied the use of neural networks to the task of style transfer. The work made use of a pretrained VGG\cite{simonyan2014vgg} to extract the semantic information from an image at different hierarchical levels. \cite{gatys2016image} optimized a loss function composed of two terms, content loss and style loss, which represented the similarity of the generated image with the content image and the style image respectively. Since \cite{gatys2016image} worked by running an optimization for each target and style image, it was not suitable for real-time applications. This was further improved by \cite{li2016precomputed, johnson2016perceptual, ulyanov2017improved, huang2017adain} by being real-time and  providing the flexibility of general style conditioning. Additionally, the works of \cite{liu2017depth, ioannou2022depth} added the sophistication of monocular depth estimation for preserving the depth information present in the source image. Furthermore, works of \cite{huang2017real,gupta2017characterizing,gao2019reconet,chen2017coherent,chen2020optical,wang2020consistent} extended the work of 2D style transfer to a video sequence by making use of advancements in methods like optical flow\cite{dosovitskiy2015flownet}. However, \cite{huang2021learning} and our experiments show that simply extending 2D style transfer to 3D scenes result in visual artifacts such as blurriness and inconsistency across different views, as also depicted in \Cref{fig:vanilla_methods}. 

% \vspace{-0.32cm}
% \subsection{Scene Stylization}
% \vspace{-0.1cm}
Stylizing scenes given a style image has gained prominence in recent years, especially with the arrival of AR/VR headsets in the mainstream.
One way to distinguish between different methods is based on their way of scene representation. Works of \cite{huang2021learning,mu20223d,cao2020psnet} work with a point cloud based representation. \cite{huang2021learning} for instance, project the style features from 2D into 3D and transform them into that of the style image before re-projecting them back into 2D for getting stylized views. Similarly, \cite{hollein2022stylemesh} performs scene stylization of indoor scenes in the form of a mesh. Recently, radiance fields \cite{mildenhall2021nerf} gained huge popularity due to their ability to generate novel views with a very high quality. Due to this, a lot of works made use of a NeRF backbone to perform scene stylization. One direction of approaching the problem is by focusing to optimize a radiance for each given style image. Methods such as \cite{jung2024geometry,zhang2022arf,li2023arfplus,chiang2022stylizing,nguyenphuoc2022snerf,Xu2023DeSRFDS,zhang2023refnpr,pang2023locally,zhang2023transforming} follow this outline. Methods such as \cite{chen2022upstnerf,chiang2022stylizing} make use of a hypernetwork to impart the style image features onto the rendered views of a complex 3D scene. On the other hand, works such as \cite{huang2022stylizednerf,liu2023stylerf} provide a generalizable scene stylization framework, i.e a stylized scene can be generated at inference time given a conditioned style image input. Our work also follows in this category. Another direction in this domain lies making use of additional information such as depth maps \cite{jung2024geometry} since their primary focus lies in geometry preservation. 
In this work, instead of a radiance field, we make use of a Gaussian splatting\cite{kerbl3Dgaussians} framework.
% To the best of our knowledge, we are the first method that builds up on a 3DGS  backbone for image-based scene stylization. 

Apart from image based conditioning for stylization, one direction is to input the conditioning in the form of text. Leveraging the use of large language models(LLMs), \cite{michel2021text2mesh,metzer2022latentnerf,hwang2023text2scene,wang2022nerfart,haque2023instruct,song2023blendingnerf,fang2023gaussianeditor} perform scene editing, and stylization in some cases, via text-based inputs.
In our work, we do not compare with any text-based input methods and focus entirely on methods that condition the scene via style images only. While textual information can be informative and descriptive, the challenges of image-based conditioning can be far more demanding as it provides information in terms of abstraction, strokes, and so on. 

%% file: sections/preliminaries.tex
\section{Preliminaries}
\label{sec:prelim}

% \vspace{-0.35cm}
\subsection{3D Gaussian Splatting}
3D Gaussian Splatting(3DGS)\cite{kerbl3Dgaussians} is an explicit method for scene representation, that is especially useful for high quality real-time rendering. We briefly introduce the main concepts for completeness.

Gaussian Splatting can be considered as a form of point cloud representation. Each scene in 3DGS is represented as a bunch of 3D Gaussians, each having certain properties. Specifically, each 3D Gaussian is defined by its mean position $\mu \in \mathbb{R}^{3}$ and a covariance matrix $\Sigma \in \mathbb{R}^{3 \times 3}$ as

\begin{equation}
\label{formula:gaussian's main formula}
    G(X)=e^{-\frac{1}{2}\mu^T\Sigma^{-1}\mu}.
\end{equation}

Since the covariance matrix has to be positive semi-definite for differentiable optimization, it is decomposed into a rotation matrix $\mathbf{R}$ and a scaling matrix $\mathbf{S} \in \mathbb{R}^{3}$ as
\begin{equation}
\label{formula:sigma decom}
    \Sigma = \mathbf{R}\mathbf{S}\mathbf{S}^T\mathbf{R}^T.
\end{equation}
Additionally, each Gaussian also contains the opacity $\alpha$ and view-dependent color values, represented by spherical harmonics coefficients. 
Once initialised, these Gaussians are projected onto 2D for rendering. This rendering process is done by a differentiable tile rasterizer. During the process of optimization, these Gaussians undergo density control, i.e they are constantly removed and new ones are added to the scene in order to maintain high image quality. This is done in order to ensure that the density of Gaussians is more or less consistent in all areas of the scene, especially ones which were empty during the SfM initialization. 
The 3D scene is optimized for the following loss function:
\begin{equation}
    \mathcal{L} = (1-\lambda)\mathcal{L}_{1} + \lambda\mathcal{L}_{D-SSIM},
\end{equation}

where $\mathcal{L}_{1}$ and $\mathcal{L}_{D-SSIM}$ are computed between the generated image and the ground truth view. For thorough explanation, we refer the reader to \cite{kerbl3Dgaussians}.

% \subsection{Multi-resolution Hash Grid}
% Encoding neural network inputs into higher dimensional space has been crucial in achieving scene representations of high quality with compact models~\cite{mildenhall2021nerf}. ~\cite{muller2022instant} furthers fidelity and efficiency by outsourcing the representation power to an auxiliary data structure, namely a multi-resolution hash grid, which allows fast access to learned features at each grid cell on multiple spatial resolutions. Specifically, an input 3D point is first parameterized by a hash code $(\boldsymbol{x}_1, \ldots, \boldsymbol{x}_l)$, which corresponds to the corner indices of a set of grids with different spatial resolution $\{V_1, \ldots, V_L\}$. Given those indices, feature vectors are obtained and linearly interpolated according to the relative position of $\boldsymbol{x}$ across all $L$ spatial resolutions. Formally, the final multi-resolution hash encoding of a 3D point $\boldsymbol{x}$ becomes $\gamma(\boldsymbol{x}) = (\gamma_1(\boldsymbol{x}_1), \ldots, \gamma_l(\boldsymbol{x}_l))$, where $\gamma_l(\boldsymbol{x}_l)$ is a feature vector obtained by means of interpolation of hash entries at the grid corners of the $l$th resolution level.

% Multi-resolution hash encoding is independent at each level and highly adaptive to the desired finest resolution. Thus, it enables faster convergence during training, does not impede inference speed, and promotes a higher level of detail, which is beneficial for the stylization task using 3DGS representation. 

\vspace{-0.3cm}
\subsection{Stylized NeRF}
StylizedNeRF \cite{huang2022stylizednerf} set up a new benchmark in the field of neural scene stylization. Built upon the framework of NeRF and Nerf-W \cite{mildenhall2021nerf, martin2021nerfw}, this work makes use of a mutual learning strategy to combine the 2D stylization method AdaIN \cite{huang2017adain} and a pre-trained NeRF architecture. The high-level idea is to predict a new stylized color conditioned on the style image. This is achieved by replacing the color component with a new style module while keeping the density prediction from the pre-trained NeRF constant to maintain geometric consistency. To facilitate mutual learning between 2D-3D stylization, StylizedNeRF makes the decoder of AdaIN trainable for fine-tuning the final results. Inspired by Nerf-W, they also employ a pre-trained VAE that provides learnable latent codes for the conditioned style inputs. Motivated by the positive results of StylizedNeRF, GSS also makes use of pre-trained Gaussians to better represent the geometric details of the complex scenes.

%% file: sections/method.tex
% \begin{minipage}{\linewidth}
% \begin{center}
% \includegraphics[width=0.9\linewidth]{images/pipeline.png}
% \captionof{figure}{PIPELINE FIGURE}
% \end{center}
% \label{fig:pipeline}
% \end{minipage}

\begin{figure}[ht!]
\centering
% \begin{subfigure}[b]{0.8\textwidth}
   \includegraphics[width=0.8\textwidth]{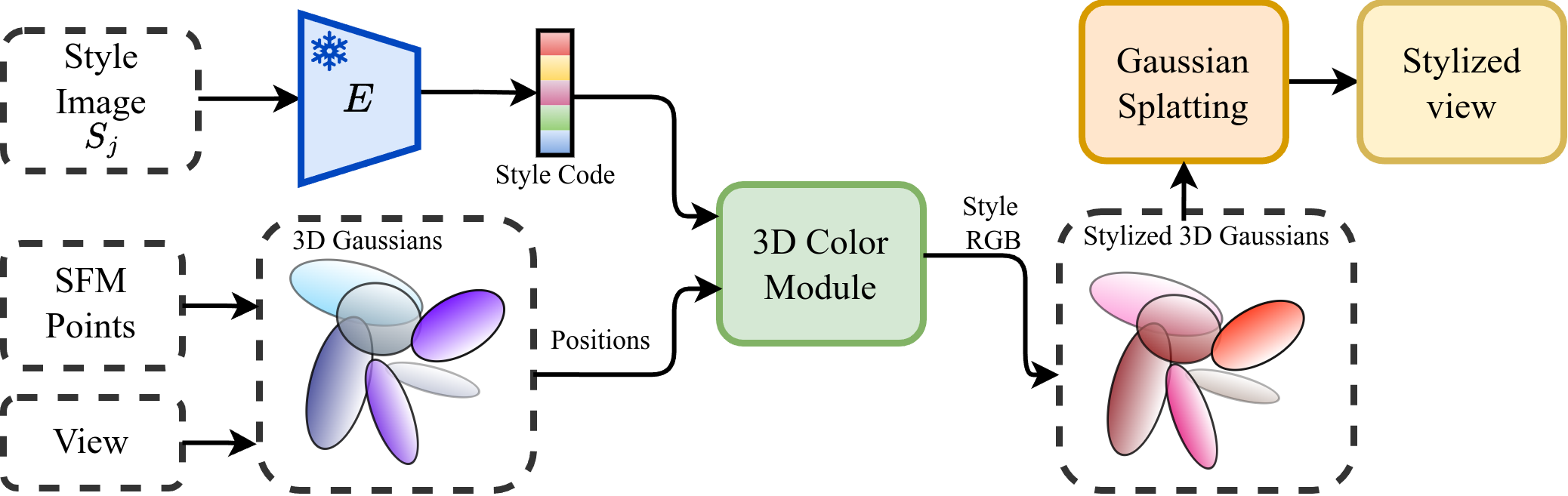}
   % \caption{}
% \end{subfigure}

% \begin{subfigure}[b]{0.8\textwidth}
%    \includegraphics[width=1\textwidth]{figures/zoomed_module-dominik.png}
%    \caption{}
   
% \end{subfigure}
\caption{Here we diagramatically show the overall architecture of our pipeline. We employ a novel 3D Color module that is jointly trained with the 3D Gaussians, to predict the new colors for each Gaussian based on the querying style image at test time.}
\vspace{-0.8cm}
\label{fig:pipeline}
\end{figure}

% \vspace{-0.5cm}
\section{Gaussian Splatting in Style}
\label{sec:method}
\vspace{-0.3cm}
Our goal is to stylize a complex 3D scene. Our overall training process is similar to that of 3DGS\cite{kerbl3Dgaussians,lee2023compact}, where the scene is learnt by minimizing the difference between multiple training views and their corresponding ground truth images. In addition to learning the scene representation, we learn a mapping between the different views of the scene and the queried style image. To this end, we employ an MLP that learns the mapping between the various 3D Gaussians and their stylized color based on the style image. During inference time, we generate novel views conditioned on an input style image. It is worth noting that the input style image at inference time may or may not have been a part of the dataset of style images used in the training process. In the following section, we elaborate on the various constituents of our pipeline. \\

% The first step is to learn the scene information from a given collection of images in a multi-view setting. We achieve this by using a 3DGS backbone. For this objective, we take pretrained models of learnt scenes for the next step. The second stage includes learning the scene information conditioned on various style images.  \\
% \linebreak
% \subsection{2D stylization network}

% \subsection{3D Color module}
\vspace{-0.3cm}
\noindent\textbf{3D Color Module} The first component of our pipeline is a 3D Color module, responsible for predicting the colors of each 3D Gaussian based on the conditioning style image. The input to this module are the mean positions $\mu$ of the 3D Gaussians and a latent code representing the style image obtained through the use of a pretrained encoder\cite{radford2021learning,
huang2022stylizednerf}. The mean input positions are fed into a multi-resolution hash grid(MHG) \cite{muller2022instant} while the latent codes are fed into an FC layer before concatenating them together. This resulting tensor is then passed through an MLP \cite{tiny-cuda-nn, lee2023compact} to obtain a new RGB color for each input Gaussian. The use of MHG and tiny MLP was motivated by the fact that each scene has a large number of Gaussians, and could also be in the order of millions for large scenes such as those in the TnT\cite{Knapitsch2017tnt} dataset.
The predicted RGB from the 3D color module is fed into the 3DGS pipeline, and the Gaussians, with their new colors are rendered to generate a 3D consistent stylised view $\mathcal{I}_{gen}$. Since our color module is view independent, it does not affect the runtime during inference time. Furthermore, the geometry is conserved, one forward pass through the proposed pipeline with the conditioned style input allows us to obtain the stylized RGB for any arbitrary viewpoint. \\

\vspace{-0.3cm}
\noindent\textbf{2D Stylization Module} The 2D Stylization moule is based on AdaIN \cite{huang2017adain}. It consists of three components, a pretrained VGG \cite{simonyan2014vgg} as the encoder, a convolutional decoder, and adaptive instance normalisation layer. For details, please refer to \cite{huang2022stylizednerf}. As shown in \Cref{fig:pipeline}, we pass the style image and the original ground truth view through a pretrained VGG to obtain their features. These features are then passed through the adaptive instance normalization layer and the decoder to obtain the 2D styled image of that particular view. This 2D stylized image acts as a guide for our 3D color module, and gives a rough indication as to how the scene should appear from that specific viewpoint. Therefore, this image, represented by $\mathcal{I}_{style2d}$ forms on what we call it as a Guide Loss $\mathcal{L}_{g}$, which is detailed in \Cref{sec:loss}. \\

\vspace{-0.3cm}
\noindent\textbf{Training Regime} We build on top of vanilla 3DGS\cite{kerbl3Dgaussians}. During the training process, we learn the scene in conjunction with the 3D Color module. To this end, we first start by learning the scene without the color module and train on unstyled training views. The motivation for this "warm-up" phase arises to encourage the network to learn the original geometry, and therefore have an inherent notion of being consistent.\\

\vspace{-0.3cm}
\noindent\textbf{Loss function}
\label{sec:loss}
As shown in \Cref{fig:pipeline}, we apply the loss functions on the views generated by the 3D color module. Our loss term is therefore made up of the \textbf{ guide loss} $\mathcal{L}_{g}$ and is elaborated as  
% \vspace{-0.2cm}
\begin{equation}
\label{loss:guide_loss}
    \mathcal{L}_{g} = ||\mathcal{I}_{style2d} - \mathcal{I}_{gen}||,
\end{equation}

% \begin{equation}
% \label{loss:content_loss}
%     \mathcal{L}_{c} = ||\mathcal{I}_{gen} - \mathcal{I}_{i}||
% \end{equation}

where $\mathcal{I}_{style2d}, \mathcal{I}_{gen} $ are the 2D stylized image from the 2D stylization module, and the stylized view from our pipeline respectively. 

%% file: sections/results.tex
\section{Experiments}
\label{sec:experiments}
% In the following section, we present the results of our method described in \Cref{sec:method}. 

\vspace{-1cm}
% \input{tables/consistency-brief}
\input{tables/consistency-brief-transposed}

In this section, we first present the implementation details of our method, followed by comparisons with existing methods and ablation studies.

\subsection{Implementation Details}
Our implementation is based on the 3DGS~\cite{kerbl3Dgaussians} framework. We closely follow the implementation of the method. We employ a warmup strategy, similar to Lee et al.~\cite{lee2023compact}, wherein we train 3DGS without our 2D stylization and the 3D Color modules. This is so that consistency is established in the initial iterations of the training and initialized Gaussians. We do this for the first 3000 iterations. Similar to 3DGS training, we allow the Gaussians to densify until 15000 iterations, after which we fix the number of Gaussians in the scene. For the 2D stylization module, we use a pre-trained VGG model~\cite{simonyan2014vgg,gatys2016image} and AdaIN decoder~\cite{huang2017adain}. We encode the input style images using a pre-trained image encoder ~\cite{radford2021learning}. We train an MLP after the warmup as described in ``training regime'' 
%\Cref{fig:pipeline}
for a total of 150k iterations. The MLP is composed of 2 fully connected layers with 64 neurons each. For the MHG, we use a resolution of 16 levels, with each level having a feature vector of length 2. For optimization, we use the Adam optimizer~ \cite{kingma2017adam} with a learning rate of $1e^{-4}$. We run all our experiments on a Nvidia A4500 GPU. \\

\subsection{3D Stylization with Gaussian Splatting}
\label{sec:baselines_datasets}
% Since no previous approach performs scene stylization based on Gaussian splatting, we compare our method with NeRF-based approaches. 

We compare our method on stylizing 3D scenes against StylizedNerf\cite{huang2022stylizednerf, hu2020jittor}, LSNV\cite{huang2021learning}, StyleRF\cite{liu2023stylerf}, and artistic radiance fields(ARF)\cite{zhang2022arf}.
We would like to emphasize that StylizedNerf, StyleRF, and LSNV are more relevant to our method as the methods focus on generalizing to multiple styles. ARF is optimized on a given style per scene. Further, we compare with two additional baselines that we call Ada-GS and GS-Ada. In Ada-GS, we train 3DGS on 2D stylized images for each style. This setting resembles that of ARF. In GS-Ada, we apply AdaIN\cite{huang2017adain} on novel views rendered from a pre-trained 3DGS model. 

\noindent\textbf{Datasets}
Following existing methods~\cite{huang2022stylizednerf, liu2023stylerf, huang2021learning}, we stylize various real-world datasets, LLFF dataset~\cite{mildenhall2019local}, and Tanks and Temples (TnT)~\cite{knapitsch2017tanks}. LLFF dataset consists of high-resolution forward-facing images of 3D scenes. We stylize \textit{trex, fern, and room} from the LLFF dataset. We learn to stylize each scene over a dataset of approximately 20000 diverse style images sampled from the WikiArt dataset~\cite{saleh2015largescale} consisting of paintings and popular artworks of various artists.

\subsubsection{Quantitative Results}
\label{sec:quantitative}
2D or 3D Stylization, in general, is a subjective task. Hence, no metrics exist that can effectively measure stylization. However, we our goal is to stylize images in a 3D consistent way. Therefore, similar to existing works~\cite{huang2021learning,huang2022stylizednerf,liu2023stylerf}, we measure short-term and long-term consistency. We use adjacent views for short-term and far-away views for long-term consistency. Similar to the evaluation in \cite{liu2023stylerf}, we first take two stylized views. By using optical flow based methods\cite{teed2020raft, niklaus2020softmax}, we then warp one view into another before computing the masked LPIPS\cite{zhang2018perceptual} and RMSE score.  Many works such as \cite{huang2021learning} also refer to it as warped LPIPS and warped RMSE. Mathematically, they can be summarised as
% $\mathbb{E}_\text{wlpips}(\mathcal{O}_{v},\mathcal{O}_{v^{'}}) = \text{LPIPS}(\mathcal{O}_{v}, \mathcal{M}_{v}(\mathcal{W}(\mathcal{O}_{v^{'}})) )$ and $\mathbb{E}_\text{wrmse}(\mathcal{O}_{v},\mathcal{O}_{v^{'}}) = \text{RMSE}(\mathcal{O}_{v}, \mathcal{M}_{v}(\mathcal{W}(\mathcal{O}_{v^{'}})) )$
\begin{equation}
    \mathbb{E}_{wlpips}(\mathcal{O}_{v},\mathcal{O}_{v^{'}}) = LPIPS(\mathcal{O}_{v}, \mathcal{M}_{v}(\mathcal{W}(\mathcal{O}_{v^{'}})) )
\end{equation}
and
\begin{equation}
    \mathbb{E}_{wrmse}(\mathcal{O}_{v},\mathcal{O}_{v^{'}}) = RMSE(\mathcal{O}_{v}, \mathcal{M}_{v}(\mathcal{W}(\mathcal{O}_{v^{'}})) ),
\end{equation}
% \begin{align}
%     \mathbb{E}_{wlpips}(\mathcal{O}_{v},\mathcal{O}_{v^{'}}) &= LPIPS(\mathcal{O}_{v}, \mathcal{M}_{v}(\mathcal{W}(\mathcal{O}_{v^{'}})) ) \\
%     \mathbb{E}_{wrmse}(\mathcal{O}_{v},\mathcal{O}_{v^{'}}) &= RMSE(\mathcal{O}_{v}, \mathcal{M}_{v}(\mathcal{W}(\mathcal{O}_{v^{'}})) ),
% \end{align}
where $\mathcal{M}_{v}$, $\mathcal{O}_{v}$, and $\mathcal{W}$ are the masking, the rendered view, and the warping function respectively for two views ${v}$ and ${v}^{'}$.

We quantitatively compare our method with the baselines as shown in \Cref{table:quantitative}. It can be clearly seen that our method outperforms all other baselines in both short-term and long-term consistency. We can see that simply applying 2D stylization to rendered novel views is not enough to ensure consistency, in accordance with \cite{huang2021learning}, and hence Ada-GS is almost always more consistent than GS-ADA for both short-term and long-term views. Furthermore, simply adding 2D stylization to the input or outputs of existing scene representation approaches is not sufficient, as all other dedicated baselines perform better on average than GS-ADA and ADA-GS. We also find that methods utilising explicit scene representations, such as GSS and LSNV\cite{huang2021learning} perform better than radiance field based baselines. It is also important to observe that for StylizedNerf, the generated novel views suffer from high blur, over-smoothing and loss of details. Due to this, they obtain a very high consistency despite not being of reasonable quality as compared to all other methods. \\

\begin{table}[h]
\centering
\caption{We perform a runtime analysis of our method vs other baselines. Due to our 3DGS backbone, we are able to generate stylized novel views approximately 4x faster than the second method, achieving an average of 157 FPS(Frames per second). Radiance fields in general have a high rendering time and suffer from slow training and testing time.}
% \vspace{-0.1cm}
\begin{tabular}{|l|l|l|l|l|l|}
\hline
                                & LSNV & ARF & StylizedNerf & StyleRF & GSS \\ \hline
Rendered FPS & 40   & 1   & 0.004        & 0.04    & \textbf{157} \\ \hline
\end{tabular}

\label{table:runtime}
% \vspace{-0.3cm}
\end{table}

% \vspace{-0.3cm}
On running on the same hardware, GSS is also the fastest. Whereas all other methods take in the order of a few seconds to minutes to generate a single view, we obtain a rendering speed of roughly averaging \textbf{157 FPS}. Our method is also efficient in training and it requires roughly 4 hours to train for each scene on a standard Nvidia A4500 GPU.

\begin{figure}[ht!]
\centering
\begin{subfigure}[b]{1.0\linewidth}
    \begin{overpic}[width=1\linewidth,tics=5, ]
        {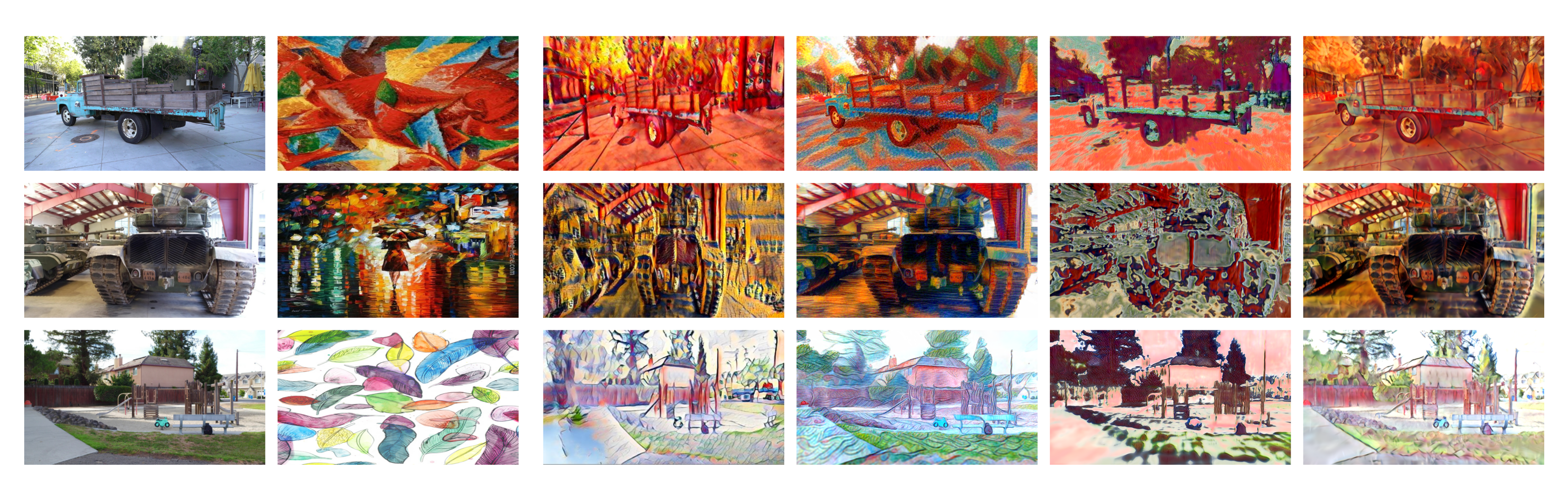}
         \put(33.8,-1){\linethickness{0.3mm}\color{black}\line(0,1){32}}
         \put(1,-1){Ground truth}
         \put(23,-1){Style}
         \put(39,-1){LSNV}
         \put(57,-1){ARF}
         \put(70.5,-1){StyleRF}
         \put(84,-1){GSS (Ours)}
    \end{overpic}
   \caption{}
\end{subfigure}

\begin{subfigure}[b]{1.0\linewidth}
   
    \begin{overpic}[width=1\linewidth,tics=5, ]{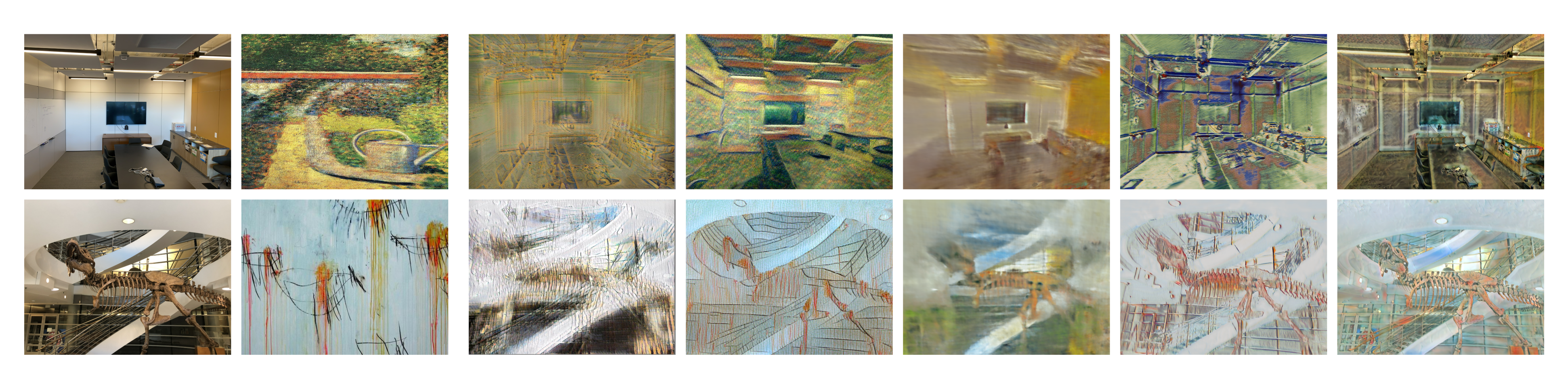}
    \put(29.25,-1){\linethickness{0.3mm}\color{black}\line(0,1){25}}
         \put(0,-1){Ground truth}
         \put(19,-1){Style}
         \put(33,-1){LSNV}
         \put(47.5,-1){ARF}
         \put(60,-1){Stylised}
         \put(61,-3.5){NeRF}
         \put(73,-1){StyleRF}
         \put(85.2,-1){GSS (Ours)}
    \end{overpic}
   \caption{}
\end{subfigure}

\caption{We provide a detailed qualitative comparison of our method against the baselines detailed in \Cref{sec:experiments} on the (a) TnT and (b) LLFF datasets. Our method achieves a highly accurate stylization based on the input style. While methods such as ARF obtain a better texture, it is attributed to the fact that it is optimized separately for each style. StylizedNerf\cite{huang2022stylizednerf} produces images that suffer from over smoothing and blurriness, while StyleRF fails to grasp the accurate style color. On the other hand, our proposed method is able to retain high details present in the unstyled view while transferring the adequate texture and colors of the style image for both, indoor and outdoor real-world datasets.}
%\vspace{-0.9cm}
\label{fig:qual_result}
\end{figure}

\subsubsection{Qualitative Results}
Along with the quantitative results, we demonstrate the performance of our method with the baselines with a qualitative evaluation as shown in \Cref{fig:qual_result}. It can be seen that not only our method preserves content details better, it also is more faithful to the style images. For instance, while StyleRF\cite{liu2023stylerf} is great at preserving geometric details, it does not adequately transfer the style image features onto the novel views. On the other hand, ARF\cite{zhang2022arf} is aggressive at transferring the style image features onto the scene, and hence often leads to a loss in the information. It also has an affinity to focus more on the central object and does not transfer the style to far-away objects in the scene. It is worth noting that ARF optimizes each scene for every style image. We also observe that StylisedNerf\cite{huang2022stylizednerf} produces blurry views while LSNV also struggles to retain fine details in the stylized views.

%% file: tables/consistency-brief-transposed.tex
% \usepackage{multirow}
\newcolumntype{R}{>{\raggedleft\arraybackslash}X}

\begin{table}[h]
\caption{Here, we demonstrate the superior performance of our method using the short-term and long-term consistency metric. We compute the metric using a VGG backbone for both, RMSE and LPIPS\cite{zhang2018perceptual}, elaborated in \Cref{sec:quantitative} . We provide a qualitative evaluation to support visual superiority of our method against the mentioned baselines in \Cref{fig:qual_result}, and also provide videos for a comprehensive evaluation in the supplementary materials.}
\begin{tabularx}{\textwidth}{@{} p{2.8cm}||R R|R R||R R|R R @{}}
% \begin{tabular}{p{2.4cm}|p{1cm} p{1cm} p{1cm} p{1cm}|p{1cm} p{1cm} p{1cm} p{1cm}}
\toprule
    % & \multicolumn{4}{c||}{\cellcolor{LimeGreen}LLFF} & \multicolumn{4}{c}{\cellcolor{Goldenrod}TnT} \\
    % & \multicolumn{2}{c}{\cellcolor{RedOrange}Short-Term} & \multicolumn{2}{c||}{\cellcolor{Cerulean}Long-Term} & \multicolumn{2}{c}{\cellcolor{RedOrange}Short-Term} & \multicolumn{2}{c}{\cellcolor{Cerulean}Long-Term} \\ % with color
    & \multicolumn{4}{c||}{LLFF\cite{mildenhall2019local}}& \multicolumn{4}{c}{TnT\cite{knapitsch2017tanks}} \\
    & \multicolumn{2}{c|}{Short-Term} & \multicolumn{2}{c||}{Long-Term} & \multicolumn{2}{c|}{Short-Term} & \multicolumn{2}{c}{Long-Term} \\ % w/o color
    \textit{Method} & \multicolumn{1}{c}{RMSE} & \multicolumn{1}{c|}{LPIPS} & \multicolumn{1}{c}{RMSE} & \multicolumn{1}{c||}{LPIPS} & \multicolumn{1}{c}{RMSE} & \multicolumn{1}{c|}{LPIPS} & \multicolumn{1}{c}{RMSE} & \multicolumn{1}{c}{LPIPS} \\
\midrule
    GS $\rightarrow$ AdaIN & 0.122 & 0.132 & 0.153 & 0.190 & 0.050 & 0.014 & 0.107 & 0.054 \\
    AdaIN $\rightarrow$ GS & 0.106 & 0.109 & 0.138 & 0.161 & \cellcolor[HTML]{FFFFC7}0.015 & \cellcolor[HTML]{FFFFC7}0.002 & \cellcolor[HTML]{FFFFC7}0.100 & 0.055 \\
    LSNV \cite{huang2021learning} & 0.069 & 0.032 & 0.070 & 0.040 & 0.030 & 0.006 & \cellcolor[HTML]{FFFFC7}0.100 & \cellcolor[HTML]{FFFFC7}0.045 \\
    ARF \cite{zhang2022arf} & 0.039 &  0.019 & 0.095 & 0.083 & \cellcolor[HTML]{9AFF99}0.012 & \cellcolor[HTML]{9AFF99}0.001 & 0.101 & 0.048 \\
    Stylised-NeRF \cite{huang2022stylizednerf} & \cellcolor[HTML]{FFFFC7}0.028 & \cellcolor[HTML]{FFFFC7}0.009 & \cellcolor[HTML]{FFFFC7}0.040 & \cellcolor[HTML]{FFFFC7}0.021 & -- & -- & -- & -- \\
    StyleRF \cite{liu2023stylerf} &  0.039 & 0.017 & 0.095 & 0.067 & 0.024 & 0.006 & 0.130 & 0.070 \\
\midrule
    Ours & \cellcolor[HTML]{9AFF99}0.019 & \cellcolor[HTML]{9AFF99}0.005 & \cellcolor[HTML]{9AFF99}0.032 & \cellcolor[HTML]{9AFF99}0.014 & \cellcolor[HTML]{FFFFC7}0.015 & \cellcolor[HTML]{FFFFC7}0.002  & \cellcolor[HTML]{9AFF99}0.083 & \cellcolor[HTML]{9AFF99}0.040 \\
\bottomrule
% \hline
% \hline
\end{tabularx}

% \vspace{-0.8cm}}
\label{table:quantitative}
\end{table}

%% file: sections/ablations.tex
\section{Ablation Studies}
\label{sec:ablations}
To study the effects of various components of our method, we perform several experiments and report their findings below.

\subsubsection{Effect of Joint Training}

% \begin{minipage}{\linewidth}
% \begin{center}
% \includegraphics[width=0.8\linewidth]{figures/joint_training.jpg}
% % \includegraphics[width=0.79\linewidth]{example-image-a}\\
% \vspace{-0.25cm}
% \captionof{figure}{We show the effect of deploying a joint-training regime of training the 3D Gaussians in conjunction with the 3D Color module. Having a joint training in an end-to-end fashion helps to preserve key details and geometry in the rendered stylized novel view.}
% \end{center}
% \label{fig:joint_training}
% % \vspace{-0.85cm}
% \end{minipage}

\begin{figure}[ht!]
\centering
% \vspace{-1cm}
% \begin{subfigure}[b]{1.0\linewidth}
    \begin{overpic}[width=0.7\linewidth,tics=5, ]
        {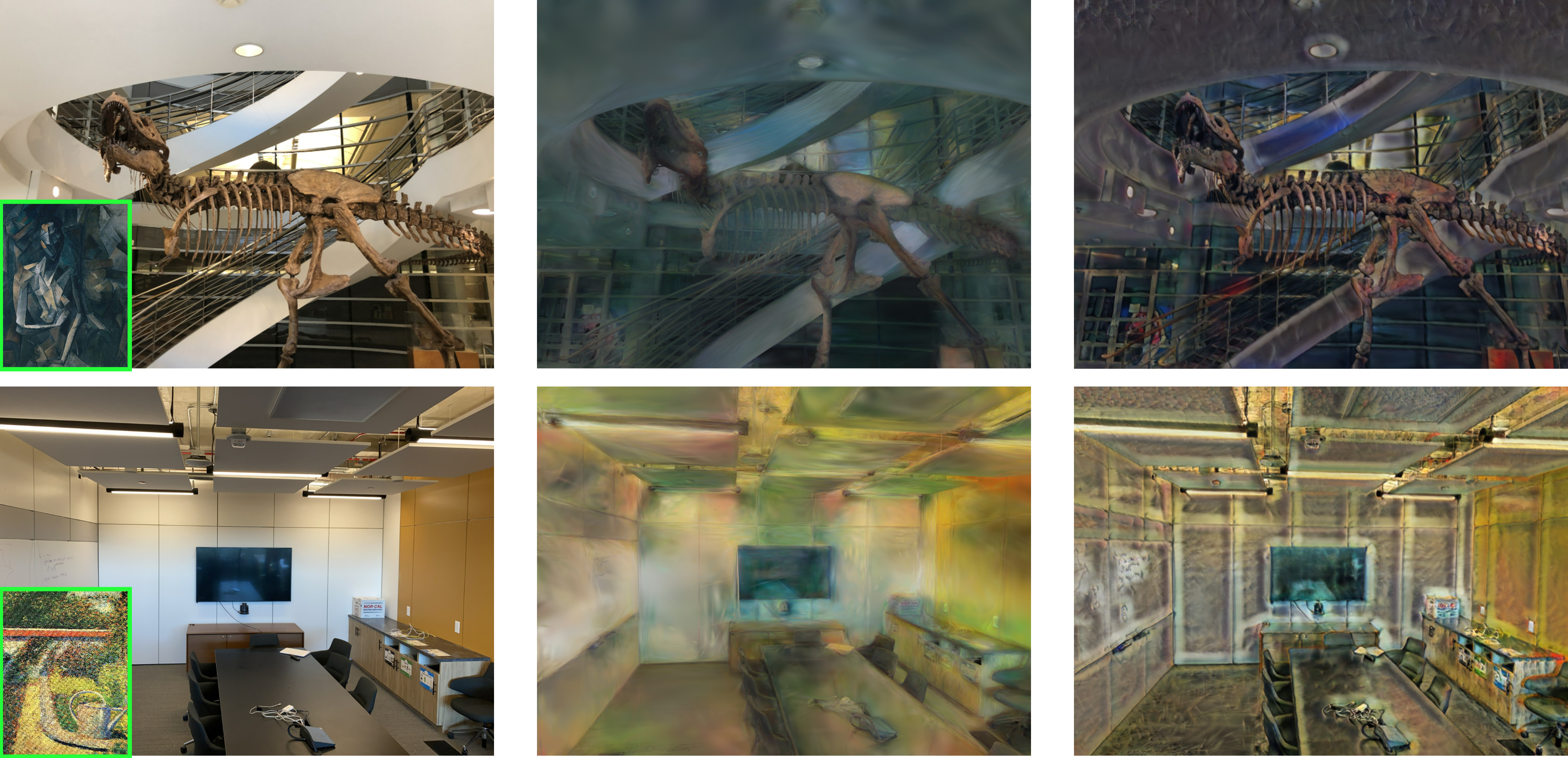}
         \put(1,-2.5){Ground truth }
         \put(1.5,-5.5){and Style}

         \put(35,-2.5){Pretrained}
         \put(35,-5.5){3D Gaussians}
         \put(76,-2.5){GSS (Ours)}
    \end{overpic}
   % \caption{}
% \end{subfigure}
\vspace{0.3cm}
\caption{We show the effect of deploying a joint-training regime of training the 3D Gaussians in conjunction with the 3D Color module. Having a joint training in an end-to-end fashion helps to preserve key details and geometry in the rendered stylized novel view.}
% \vspace{-0.5cm}
\label{fig:joint_training}
\end{figure}

Contrary to previous NeRF-based baselines such as StylizedNerf\cite{huang2022stylizednerf}, we employ a joint-training regime where the 3D Gaussians of each scene are trained in conjuction with the 3D color module. To study the effect of this regime, we perform an ablation where we employ pre-trained 3D Gaussians. These Gaussians were trained on unstylized images, and were fixed throughout the training process while the 3D Color module was trained from scratch. The stylization results from pretrained Gaussians appear blurry while our method retains significantly higher amount of details. Furthermore, since the geometric parameters such as opacity are fixed in the pretrained model, the final style imparted onto the rendered view lacks coherence.

\subsubsection{Train on stylized images vs stylizing novel renderings}

\begin{figure}[ht!]
\vspace{-0.5cm}
\centering
    \begin{overpic}[width=0.8\linewidth,tics=5, ]
        {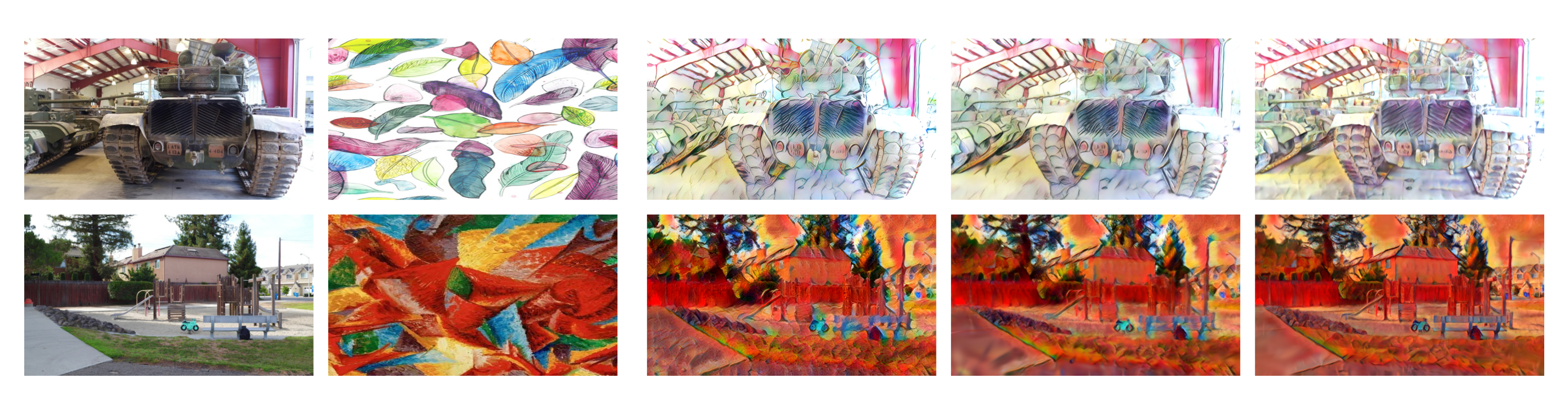}
    \put(40.25,-1){\linethickness{0.3mm}\color{black}\line(0,1){25}}
         \put(3,-1){Ground}
         \put(3, -3.5){truth}
         \put(28,-1){Style}
         \put(45,-1){GS-Ada}
         \put(65.5,-1){Ada-GS}
         \put(83,-1){GSS (Ours)}
    \end{overpic}

\vspace{0.3cm}
\caption{In addition to providing quantitative results in \Cref{table:quantitative}, we provide the qualitative results showing the distored geometry when we apply 2D stylization on rendered novel views(GS-Ada). We also observe that training vanilla 3DGS on 2D Stylized views(Ada-GS) preserves better details than GS-Ada. However, our method provides the most geometrically accurate renderings while being truthful to the queried style.}
\label{fig:vanilla_methods}
\vspace{-0.3cm}
\end{figure}

Additionally, we quantitatively and qualitatively evaluate the performance of our synthetic baselines, namely GS-Ada and Ada-GS in \Cref{fig:vanilla_methods,table:quantitative}. We elaborate on these two baselines in \Cref{sec:baselines_datasets}. We observe that GS-Ada is the least consistent as the rendered novel views are stylized independently of each other using AdaIN \cite{huang2017adain}, therefore, have no notion of consistency. While Ada-GS performs better, it has a major drawback, i.e. the method needs to be trained from scratch for every new style image, thus limiting its usability, especially for practical applications.

\begin{figure}[ht!]
% \vspace{-0.8cm}
\begin{center}
\includegraphics[width=1\linewidth]{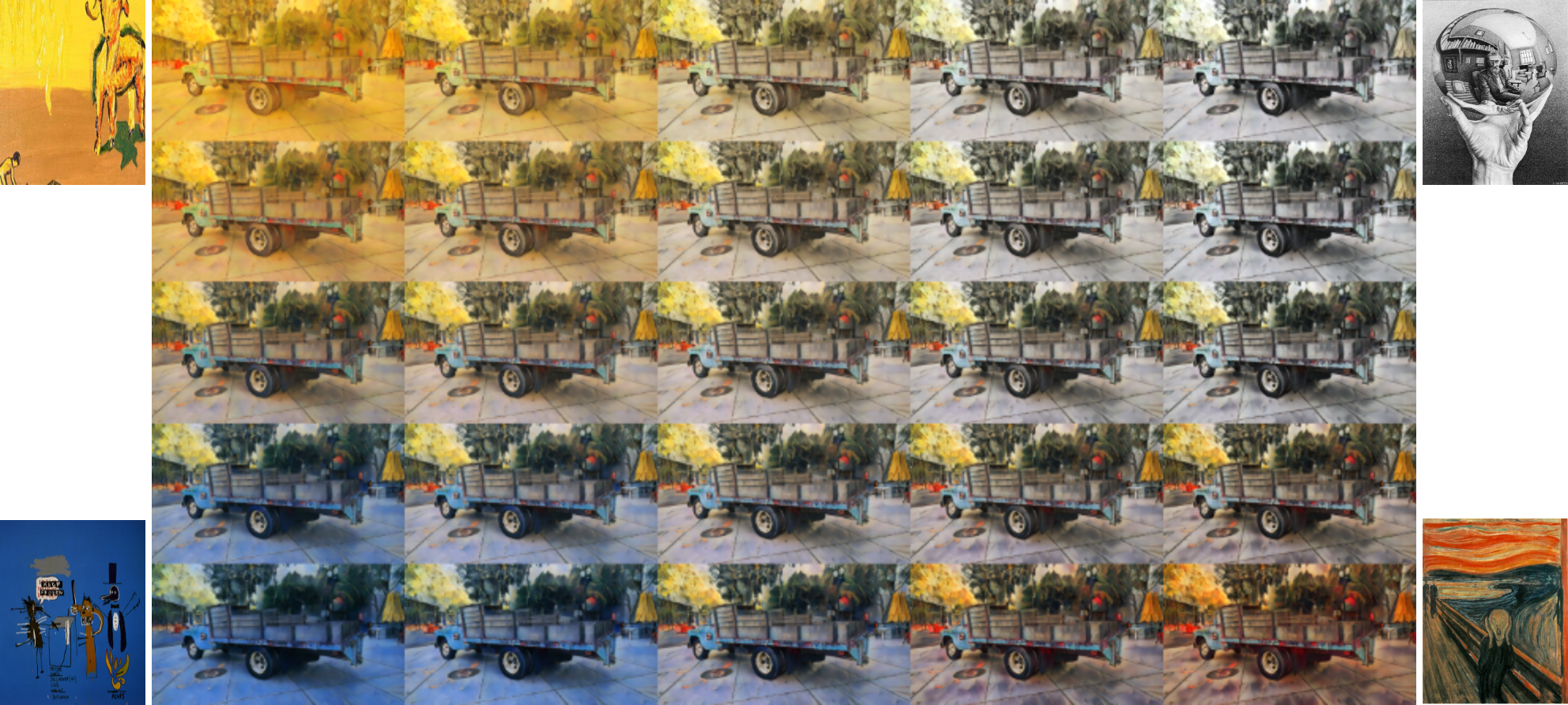}
\captionof{figure}{We interpolate between the latent vectors of the style images. The four style images are shown at the four corners of the image above and were chosen so they're highly diverse and accurately depict the differences in the predicted RGBs for each Gaussian from our model.}
\end{center}
\label{fig:interpolation-ablation}
\vspace{-1cm}
\end{figure}

\subsubsection{Style Interpolation}
% \vspace{-0.2cm}
We perform style interpolation by taking four different style images at test time, and obtaining the resulting style latent by linearly interpolating between the different styles. We show the results in \Cref{fig:interpolation-ablation}. The ablation shows the capability of our model to generate results for not only the given style images, but also a mixture of such styles. It demonstrates that our model learns the mapping from style inputs to predicted RGB values for each 3D Gaussian in a meaningful way, and thus provides increased generalisability. 

%% file: sections/conclusion.tex
\section{Conclusion}
\vspace{-0.4cm}
In this paper, we presented a novel method to stylize complex 3D scenes that are spatially consistent. Contrary to a majority of existing works, once trained, our method is capable to take unseen input scenes at inference time and produce novel views in real-time. By leveraging a multi-resolution hash grid and a tiny MLP, we are able to accurately generate the stylized colors of each 3D Gaussian present in a scene. Since we only do one forward pass through the 3D color module, we are able to generate novel views at around \textbf{150FPS}. We exhibit that GSS produces superior results by the use of quantitative and qualitative results, thus making GSS suitable for many practical applications. 